\documentclass[runningheads]{llncs}
\usepackage[T1]{fontenc}
\usepackage{graphicx,verbatim}
\usepackage{epsfig}
\usepackage{xcolor, soul}
\sethlcolor{red}
\usepackage{boldline}
\usepackage{amsmath}
\usepackage{amssymb}
\usepackage{soul}
\usepackage{url}
\usepackage{booktabs}
\usepackage{comment}
\usepackage{nccmath}
\usepackage{mathtools, cases}
\usepackage{color}
\usepackage{lipsum}
\usepackage{float}
\usepackage{multirow}
\usepackage{xcolor}
\usepackage{hhline,colortbl}
\usepackage{ctable}
\usepackage{hyperref}
\usepackage{subcaption}
\usepackage[normalem]{ulem}

\usepackage{amssymb}
\usepackage{pifont}
\definecolor{awesomePINK}{rgb}{1.0, 0.13, 0.32}
\definecolor{DarkGreen}{RGB}{1,50,32}
\definecolor{awesomeGRAY}{rgb}{0.5,0.5,0.5}
\definecolor{awesomeYELLOW}{rgb}{0.99, 0.93, 0.0}
\definecolor{TableYELLOW}{rgb}{0.98, 0.91, 0.71}

\definecolor{Gray}{gray}{0.85}
\definecolor{LightCyan}{rgb}{0.88,1,1}

\usepackage{blindtext}
\usepackage[export]{adjustbox}
\usepackage{wrapfig}
\usepackage{adjustbox}

\usepackage{xcolor}
\usepackage{color}
\begin{document}
\title{Induce to Empower: Improving Lightweight Baselines via Foundation Model Induction for Generalized Polyp Segmentation}
%

\author{
Shivanshu Agnihotri\inst{1}
\and Snehashis Majhi\inst{2}
\and Deepak Ranjan Nayak\inst{1}
\and Dwarikanath Mahapatra\inst{3}
\and Debesh Jha\inst{4}
}

\authorrunning{Agnihotri et al.}

\institute{
Malaviya National Institute of Technology Jaipur, India
\and
Côte d’Azur University, France\\
\and
Khalifa University, UAE\\
\and
Biomedical Perception \& Intelligence Lab, University of South Dakota, USA\\
\email{\textcolor{blue}{drnayak.cse@mnit.ac.in}}\\
\url{https://github.com/lostinrepo/Lite-Pi}
}
  
\maketitle              
\begin{abstract}
Automated polyp segmentation in colonoscopy continues to pose challenges due to substantial appearance variations and indistinct polyp boundaries. Although emerging foundation models (FMs) such as DINOv2, SAM, and OneFormer, demonstrate remarkable generalization capabilities, their direct transfer to the polyp segmentation task and deployment in real-time clinical settings are difficult due to lack of large-scale labeled data and high computational demands. In addition, adopting multiple FMs together raises concerns, even though they encode complementary semantic and structural information.  While lightweight models, including U-Net, PraNet and U-Net++, are computationally efficient, they often struggle to generalize across datasets due to limited representational capacity. 
To address this gap, we propose \textbf{\uline{Lite}}-\textbf{\uline{P}}olyp\textbf{\uline{I}}nductor (Lite-$\pi$), a novel foundation model induction framework that significantly enhances lightweight polyp segmentation baselines. Our proposed framework generates FM-specific prototype representations and aligns them semantically with the corresponding foundation model priors through reconstruction-based supervision. Subsequently, transformer-based fusion is introduced to highlight the polyp-relevant representations, including salient boundary information, while preserving complementary semantic cues.
Extensive experiments across five polyp segmentation benchmark datasets demonstrate that Lite-$\pi$ significantly improves lightweight baselines, achieving superior generalization performance with minimal computational overhead and thereby, offering a practical solution for generalized polyp segmentation. Our code is available at \href{https://github.com/lostinrepo/Lite-Pi}{GitHub}.

\end{abstract}
\section{Introduction}
Colorectal cancer (CRC) is a leading cause of cancer-related mortality worldwide, with its incidence increasing across diverse age groups. Most CRC cases originate from benign colorectal polyps, making early detection and subsequent removal crucial for effective prevention. While colonoscopy is considered the gold standard for CRC screening, it is highly operator-dependent and subject to variability among clinicians. Further, the high polyp miss rates of 6–27\% \cite{ahn2012miss} during routine colonoscopy underscore the pressing need for computer-aided polyp segmentation approaches to aid clinicians in accurate detection and timely intervention.

Deep learning–based automated polyp segmentation has gained significant attention and has shown promising results in recent years. Specifically, encoder- decoder Convolutional Neural Network (CNN) architectures, including U-Net and its variants \cite{ronneberger2015u}  \cite{zhou2018unet++} \cite{huang2020unet} have been widely adopted, but often struggle to preserve fine boundary details. Subsequently, various models such as PraNet \cite{fan2020pranet}, SFA \cite{fang2019selective}, MSNet \cite{zhao2021automatic}, M$^2$SNet \cite{zhao2023m}, and CFA-Net \cite{zhou2023cross}, have been proposed to address boundary-related issues and scale variations in polyps. However, these methods fall short of modeling global contextual details, resulting in suboptimal performance. On the other hand, Vision Transformer (ViT) \cite{dosovitskiy2021an} and hybrid CNN-ViT models such as PVT-Cascade \cite{rahman2023medical}, Polyp-PVT \cite{Dong2023}, and  CTNet \cite{xiao2024ctnet} excel at modeling global and/or local relationships, yielding rich feature representations and significantly improving polyp segmentation accuracy. Despite strong performance improvements, high computational demands and limited generalization hinder their clinical applicability. Additionally, achieving robust polyp segmentation still remains challenging due to large variations in polyp appearance, ambiguous boundaries and cross-dataset domain shifts.

Foundation Models (FMs), such as SAM \cite{kirillov2023segment}, DINOv2 \cite{oquab2024dinov}, OneFormer \cite{jain2023oneformer}, CLIP \cite{radford2021learning}, MaskFormer \cite{cheng2021per} and Mask2Former \cite{cheng2022masked}, have significantly advanced image segmentation by learning rich visual representations from large-scale data that facilitates strong cross-domain generalization. However, their direct application to polyp segmentation is challenging due to data scarcity, high computational overhead, increased memory demands, and insufficient domain-specific knowledge. To mitigate these issues, SAM-Mamba \cite{dutta2025sam} has been proposed, introducing adapter-based tuning via a Mamba-Prior module to improve generalization performance. Following this, SAM-MaGuP \cite{dutta2025ma} has recently been introduced, incorporating a Mamba-guided boundary prior along with a 1D–2D Mamba adapter to address the weak-boundary challenge. However, these approaches incur substantial computational overhead, requiring approximately 103–106M parameters and 423–431 GFLOPs, which limits their real-time clinical deployment. 
\begin{figure*}[t]
  \centering
   \includegraphics[height=4.75cm, width=\linewidth]{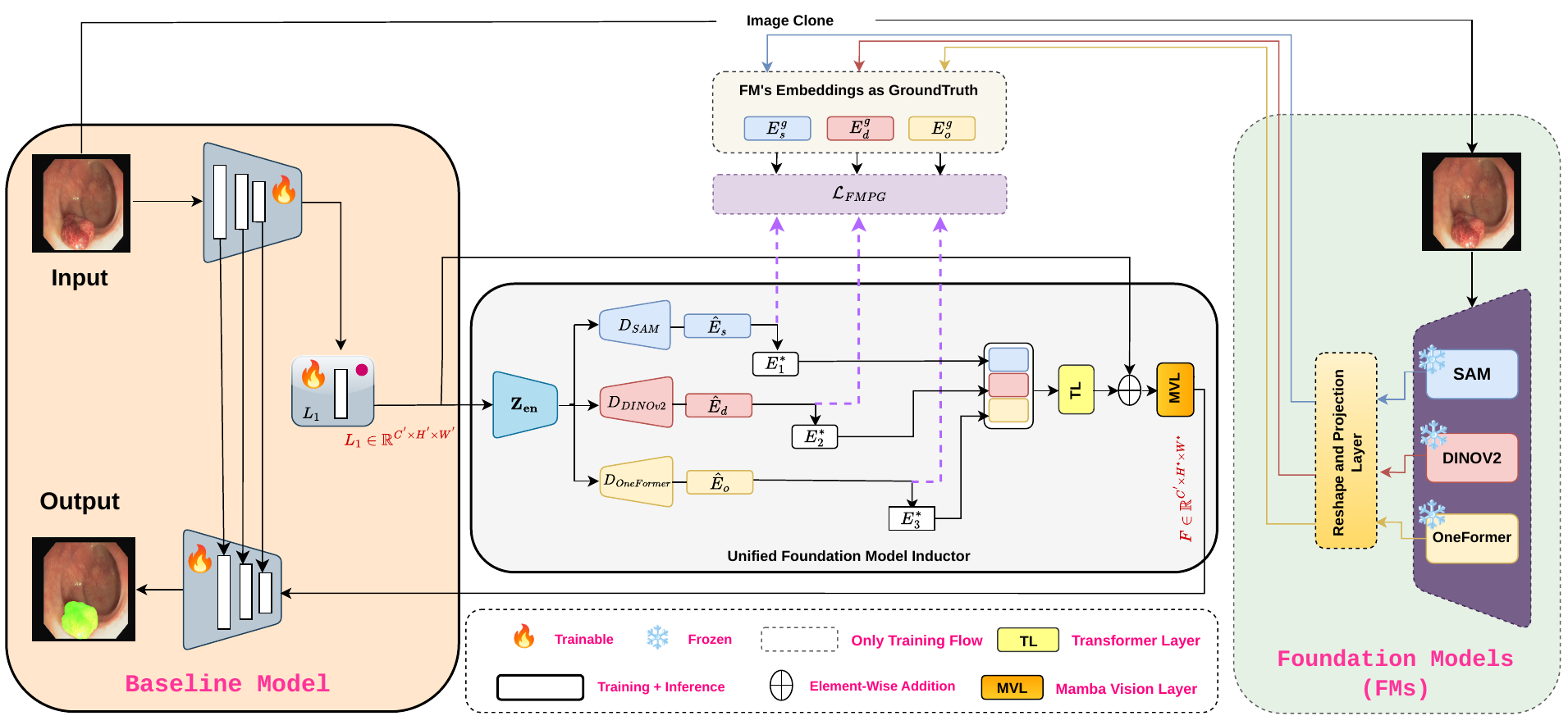}
   \vspace{-0.5cm}
   \caption{Overall pipeline of the proposed \textbf{Lite-$\pi$} framework. It incorporates a novel inductor module which generates synthetic prototype embeddings and performs semantic alignment with structural priors from foundation models to enhance representation capability of lightweight segmentation baselines.} 


   \label{fig_model}
\end{figure*}
Polyp-DiFoM~\cite{agnihotri2026sam}, a distillation-based framework, mitigates deployment challenges by transferring enriched representations from multiple FMs into lightweight segmentation baselines. Although this framework significantly improves the performance of lightweight baselines, their performance and generalizability still fall short of meeting real-time clinical demands. {This limitation stems from representational redundancies and task-bias introduced when leveraging multiple FMs.


Motivated by the above analysis, we propose \textbf{\uline{Lite}}-\textbf{\uline{P}}olyp\textbf{\uline{I}}nductor (Lite-$\pi$), a novel framework, that introduces \textbf{Unified Foundation Model Inductor (UFMI)} module to improve the performance of lightweight segmentation baselines  (U-Net, U-Net++, and PraNet) with minimal additional computational overhead. Unlike conventional knowledge transfer approaches, Lite-$\pi$ induces FM-specific synthetic prototype embeddings from the intermediate representations of a lightweight network and semantically aligns them with structural priors derived from corresponding FMs. Further, it selectively amplifies polyp-relevant representations, enabling enhanced global contextual dependency modeling and fine-grained feature refinement, thereby improving segmentation performance.
Our framework is evaluated on five benchmark datasets, demonstrating significant performance gains and improved generalization over lightweight models, achieving near state-of-the-art (SoTA) performance. 

Our \textbf{contributions} are three-fold:  (1) We introduce {Lite-$\pi$} that leverages three FMs (SAM, DINOv2, and OneFormer) to inject task-specific and rich semantic representations into lightweight models for improved polyp segmentation. (2) We propose a novel inductor module, which generates prototype representations from the lightweight encoder and enforces semantic alignment with structural priors extracted from three FMs via a reconstruction loss, enriching feature representations of lightweight models while ensuring their alignment is relevant to polyp segmentation. (3) We demonstrate consistent performance improvements across five benchmark datasets with strong generalization capabilities with low computational overhead, making it well-suited for real-world clinical applications.

\section{Methodology}
Our proposed Lite-$\pi$ framework consists of three key components: a  \textbf{Baseline-Specific Encoder}, a \textbf{Unified Foundation Model Inductor (UFMI)}, and a \textbf{Foundation Model Induced Decoder}, as illustrated in Figure \ref{fig_model}. The Lite-$\pi$ aims to effectively transfer crucial visual semantics and relevant task-specific information from three powerful FMs into lightweight models.

\subsection{Baseline-Specific Encoder}

Let \(\mathbf{I} \in \mathbb{R}^{H \times W \times 3}\) be the input colonoscopy image, which is fed to a lightweight encoder (U-Net, U-Net++, PraNet) consisting of a sequence of convolution layers, resulting in hierarchical feature maps \(\{{f}_1, {f}_2, \ldots, {f}_n\}\). To extract a latent representation (${L}_1$), we apply $1 \times 1$ convolution over the deepest feature map (${f}_n$) of the encoder. \vspace{-0.35cm}
\begin{equation}
{L}_1 = \text{Conv}_{1 \times 1}({f}_n) \in \mathbb{R}^{C^{'} \times {H}^{'} \times {W}^{'}}
\end{equation}
These representations (${L}_{1}$) serve as the base for the foundation model induction process. 

\subsection{Unified Foundation Model Inductor (UFMI)}
The key innovation of our approach lies in the UFMI module, which refines the latent feature representations of lightweight baselines by generating FM prototypes relevant to polyps and performing semantic alignment with structural priors from foundation models (SAM, DINOv2 and OneFormer), thereby improving polyp segmentation. 

\subsubsection{Foundation Model Prototype Generation (FMPG)}
It generates prototype embeddings from ${L}_{1}$ corresponding to each foundation model: \(\hat{{E}}_{\text{s}} \in \mathbb{R}^{C^{*} \times H^{*} \times W^{*}}\) for SAM, \(\hat{{E}}_{\text{d}} \in \mathbb{R}^{C^{*} \times H^{*} \times W^{*}}\) for DINOv2 and \(\hat{{E}}_{\text{o}} \in \mathbb{R}^{C^{*} \times H^{*} \times W^{*}}\) for OneFormer. This approach is introduced to selectively learn crucial priors essential for polyp segmentation while suppressing noise and redundancies from original FM embeddings. To achieve this, we adopt a lightweight encoder–decoder framework consisting of a single encoder followed by three parallel decoders. The encoder $\mathbf{Z_{en}} $ uses two \(3\times3\) convolutional layers to project the latent embeddings to a low-dimensional latent space. While the decoder $\text{$D_{i}$}(.)$ employs a \(3\times3\) convolution followed by a \(1\times1\) convolution to project the resultant embeddings to FM-specific representations, while preserving the dimensionality. Finally, a \(1\times1\) convolution is applied to match the dimensionality of the ground truth-embeddings generated from foundation models. 
\begin{equation}
\mathbf{Z_{en}} = Conv_{3 \times 3}(Conv_{3 \times 3}(\mathbf{L}_1))
\end{equation}

\begin{equation}
\hat{E}_{\text{i}} = Conv_{1 \times 1}(\text{$D_{i}$}(\mathbf{Z_{en}})), i \in \{s,d,o\} 
\end{equation}


\subsubsection{FM-based Semantic Feature Extraction}
We leverage three powerful FMs (SAM, DINOv2 and OneFormer) to extract semantically enriched feature embeddings $e_i \in \mathbb{R}^{c \times h \times w }$  from the input image $I$. These  embeddings are then rescaled using bilinear interpolation followed by a \(1\times1\) convolution to generate ground-truth embeddings $E^{g}_{i} \in \mathbb{R}^{C^{*} \times H^{*} \times W^{*} }$ while ensuring consistent spatial dimensions with $\hat{E}_{\text{i}}$. 

\subsubsection{Training Objective for FMPG}
Training of FMPG ensures better generation of synthetic prototypes specific to polyp characteristics. This is achieved through 
feature alignment between reconstructed embeddings (\(\hat{{E}}_{\text{s}}, \hat{{E}}_{\text{d}}, \hat{{E}}_{\text{o}}\)) and their corresponding ground-truth embeddings $({{E}^{g}_{s}}, {{E}^{g}_{d}}, {{E}^{g}_{o}}$) generated from FMs. The training objective for this stage is defined as follows.

\begin{equation}
\mathcal{L}_{\text{FMPG}} =
\sum_{i \in \{\text{s}, \text{d}, \text{o}\}}
\frac{1}{C^{*}}
\sum_{k=1}^{C^{*}}
\left( \hat{E}_{i} - E_{i}^{g} \right)^2
\end{equation}

\subsubsection{Transformer and Mamba-based Fusion}
To emphasize the polyp-relevant features and extract complementary features from FMs, we first concatenate the aligned embeddings $(E^{*}_{1}, E^{*}_{2}, E^{*}_{3})$ and subsequently employ a $1\times1$ convolution to obtain a compact representation $\mathbf{x}$. Next, $\mathbf{x}$ is processed through a transformer layer $TL(.)$, which employs self-attention to focus the most informative representations of FMs by modeling cross-correlations among them. To preserve fine-grained structural information, a skip connection is incorporated from the latent representation ${L}_1$. Finally, the representations are fed to a Mamba Vision Layer (MVL)~\cite{Hatamizadeh_2025_CVPR} to further enhance the modeling of global contextual relationships with linear complexity, critical for precise polyp segmentation. 

\begin{equation}
{F} =
\text{MVL} \left(
\text{TL}(\mathbf{x}) +
{L}_{1}
\right)
\end{equation}

\subsection{Foundation Model Induced Decoder}
The refined representation ${F}$, enriched with crucial priors from multiple FMs, including boundary cues and semantic awareness, is forwarded to the decoder of the lightweight baseline. The decoder receives two feature streams: (i) the hierarchical encoder feature maps and (ii) the refined foundation-induced embeddings. The decoding pathway follows the native design of each backbone (e.g., progressive upsampling in U-Net), and the final segmentation map is produced using a convolutional head.
\subsection{Final Training Objective}
\label{sec:training_strategy}
We start by training the baseline model using a combined Binary Cross-Entropy (BCE) and Dice loss for stable segmentation and then introduce a progressive weighted training strategy that effectively balances segmentation supervision with FMPG loss. The total training loss $\mathcal{L}_{\text{total}}$ is defined as:
\begin{equation}
\mathcal{L}_{\text{total}} = \mathcal{L}_{\text{seg}} + \lambda_p \cdot \mathcal{L}_{\text{FMPG}},
\end{equation}
\begin{equation}
\mathcal{L}_{\text{seg}} = \mathcal{L}_{\text{dice}} + \mathcal{L}_{\text{bce}},
\end{equation}

where, $\lambda_p$ indicates the weight factor, which is progressively increased during training (0.1 for first 30 epochs, 0.3 from 30 to 60 epochs, and 0.5 in later epochs), allowing the model to leverage rich prior knowledge while progressively refining its spatial representations.

\begin{table*}[t]
\centering
\caption{Quantitative comparison of baselines with Lite-$\pi$ against SoTA methods on seen datasets.}
\label{tab:combined_seen}

\footnotesize
\renewcommand{\arraystretch}{0.85}
\setlength{\tabcolsep}{3pt}
\resizebox{\textwidth}{!}{%
\begin{tabular}{lcc|cccccc|ccccccc}
\toprule
\toprule
\multirow{2}{*}{\textbf{Methods}} & \textbf{Params} & \textbf{FLOPs}
& \multicolumn{6}{c|}{\textbf{Kvasir-SEG (Seen)}}
& \multicolumn{6}{c}{\textbf{CVC-ClinicDB (Seen)}} \\
\cmidrule(lr){4-9} \cmidrule(lr){10-15}
& \textbf{(M)} & \textbf{(G)}
& mDice$\uparrow$ & mIoU$\uparrow$ & $F_{\beta}^{w}$$\uparrow$ & $S_{\alpha}$$\uparrow$ & $E_{\phi}^{\max}$$\uparrow$ & $\mathcal{M}$$\downarrow$
& mDice$\uparrow$ & mIoU$\uparrow$ & $F_{\beta}^{w}$$\uparrow$ & $S_{\alpha}$$\uparrow$ & $E_{\phi}^{\max}$$\uparrow$ & $\mathcal{M}$$\downarrow$ \\
\midrule

\multicolumn{15}{c}{\textcolor{gray}{\textit{\textbf{State-of-the-art methods}}}} \\
SANet \cite{wei2021shallow}  \textcolor{gray}{(MICCAI'21)}      & 23.8 & 11.3 & 90.4 & 84.7 & 89.2 & 91.5 & 95.3 & 2.8 & 91.6 & 85.9 & 90.9 & 93.9 & 97.6 & 1.2 \\
MSNet \cite{zhao2021automatic} \textcolor{gray}{(MICCAI'21)}    & 27.6 & 17.0 & 90.7 & 86.2 & 89.3 & 92.2 & 94.4 & 2.8 & 92.1 & 87.9 & 91.4 & 94.1 & 97.2 & 0.8 \\
Polyp-PVT \cite{Dong2023}  \textcolor{gray}{(CAAI'23)}        & 25.1 & 10.1 & 91.7 & 86.4 & 91.1 & 92.5 & 95.6 & 2.3 & 93.7 & 88.9 & 93.6 & 94.9 & 98.5 & 0.6 \\
PVT-Cascade \cite{rahman2023medical} \textcolor{gray}{(WACV'23)} & 35.2 & 32.5 & 91.1 & 86.3 & 90.6 & 91.9 & 96.1 & 2.5 & 91.9 & 87.2 & 91.8 & 93.6 & 96.9 & 1.3 \\
CFA-Net \cite{zhou2023cross} \textcolor{gray}{(PR'23)}    & 25.2 & 55.3 & 91.5 & 86.1 & 90.3 & 92.4 & 96.2 & 2.3 & 93.3 & 88.3 & 92.4 & 95.0 & 98.9 & 0.7 \\
CTNet \cite{xiao2024ctnet} \textcolor{gray}{(TCYB'24)}         & 44.2 & 32.6 & 91.7 & 86.3 & 91.0 & 92.8 & 95.9 & 2.3 & 93.6 & 88.7 & 93.4 & 95.2 & 98.3 & 0.6 \\
MEGANet \cite{bui2024meganet}   \textcolor{gray}{(WACV'24)}    & 44.1 & 28.8 & 91.3 & 86.3 & 90.7 & 91.8 & 95.9 & 2.5 & 93.8 & 89.4 & 94.0 & 95.0 & 98.6 & 0.6 \\

SAM-Mamba \cite{dutta2025sam}   \textcolor{gray}{(WACV'24)}    & 103.0 & 423.0 & 92.4 & 87.3 & 94.2 & 93.6 & 96.1 & 2.5 & 94.2 & 88.7 & 94.3 & 95.5 & 98.2 & 0.6 \\
SAM-MaGuP \cite{dutta2025ma}    \textcolor{gray}{(MICCAI'25)} & 106.0 & 431.0   & 94.7 & 89.0 & 95.1 & 94.2 & 98.1 & 1.6 & 95.3 & 91.3 & 95.8 & 96.3 & 98.8 &0.5 \\
\midrule

\multicolumn{15}{c}{\textcolor{gray}{\textit{\textbf{Lightweight baselines}}}} \\
U-Net \cite{ronneberger2015u} \textcolor{gray}{(MICCAI'15)}      & 16.7 & 73.9 & 81.8 & 74.6 & 79.4 & 85.8 & 89.3 & 5.5 & 82.3 & 75.5 & 81.1 & 88.9 & 95.4 & 1.9 \\
U-Net++ \cite{zhou2018unet++}   \textcolor{gray}{(DLMIA'18)}    & 9.1  & 65.9 & 82.1 & 74.3 & 80.8 & 86.2 & 91.0 & 4.8 & 79.4 & 72.9 & 78.5 & 87.3 & 93.1 & 2.2 \\
PraNet \cite{fan2020pranet}   \textcolor{gray}{(MICCAI'20)}      & 30.4 & 13.1 & 89.8 & 84.0 & 88.5 & 91.5 & 94.8 & 3.0 & 89.9 & 84.9 & 89.6 & 93.6 & 97.9 & 0.9 \\
\midrule

\multicolumn{15}{c}{\textcolor{gray}{\textit{\textbf{Lightweight baselines with Polyp-DiFoM (WACV'26)  \cite{agnihotri2026sam}}}}} \\
Polyp-DiFoM (U-Net)     & 16.8 & 74.7 & 86.6 & 76.4 & 85.3 & 94.1 & 91.0 & 4.1 & 93.9 & 88.7 & 92.7 & 96.9 & 96.3 & 1.1 \\
Polyp-DiFoM (U-Net++)      & 9.6  & 66.4 & 84.7 & 74.7 & 83.3 & 93.7 & 92.9 & 4.8 & 89.5 & 83.4 & 91.1 & 96.7 & 95.4 & 1.9 \\
Polyp-DiFoM (PraNet)       & 31.4 & 13.5 & 90.9 & 84.7 & 88.2 & 94.6 & 91.9 & 3.4 & 94.2 & 91.2 & 92.9 & 99.1 & 98.0 & 1.4 \\
\bottomrule
\multicolumn{15}{c}{\textcolor{gray}{\textit{\textbf{Lightweight baselines empowered with Lite-$\pi$  (Ours)}}}} \\
Lite-$\pi$ (U-Net)      & 19.6 & 74.3  & 87.4  & 80.2  & 86.1  & 92.1  & 91.4  & 4.0  &  94.9  & 90.1  & 94.5  & 96.6  & 98.9  & 0.9    \\
& &  &
\textcolor{blue}{(+5.6)} & \textcolor{blue}{(+5.6)} &
\textcolor{blue}{(+6.7)} & \textcolor{blue}{(+6.3)} &
\textcolor{blue}{(+2.1)} & \textcolor{red}{(-1.5)} &
\textcolor{blue}{(+12.6)} & \textcolor{blue}{(+14.6)} &
\textcolor{blue}{(+13.4)} & \textcolor{blue}{(+7.7)} &
\textcolor{blue}{(+3.5)} & \textcolor{red}{(-1.0)} \\
Lite-$\pi$ (U-Net++)     & 10.1  & 66.6 & 87.6 & 81.5  & 87.3  & 93.9 & 92.4 & 3.8 & 93.3  & 87.7  & 93.1 & 96.7  & 90.7  & 1.0  \\
& &  &
\textcolor{blue}{(+5.5)} & \textcolor{blue}{(+7.2)} &
\textcolor{blue}{(+6.5)} & \textcolor{blue}{(+7.7)} &
\textcolor{blue}{(+1.4)} & \textcolor{red}{(-1.0)} &
\textcolor{blue}{(+13.9)} & \textcolor{blue}{(+14.8)} &
\textcolor{blue}{(+14.6)} & \textcolor{blue}{(+9.4)} &
(-3.4) & \textcolor{red}{(-1.2)} \\
Lite-$\pi$ (PraNet)     & 34.9 & 13.9 & 92.9  & 87.7  & 88.5  & 93.7  & 93.1  & 2.7   & 95.4  & 91.7 & 94.9  & 97.8 & 99.1  & 0.8 \\
& &  &
\textcolor{blue}{(+3.1)} & \textcolor{blue}{(+3.7)} &
(0.0) & \textcolor{blue}{(+2.2)} &
(-1.7) & \textcolor{red}{(-0.3)} &
\textcolor{blue}{(+5.5)} & \textcolor{blue}{(+6.8)} &
\textcolor{blue}{(+5.3)} & \textcolor{blue}{(+4.2)} &
\textcolor{blue}{(+1.2)} & \textcolor{red}{(-0.1)} \\
\bottomrule
\bottomrule
\end{tabular}
}\vspace{-0.5cm}
\end{table*}

\section{Experiment and Results}
\textbf{Datasets:} We evaluate the effectiveness of our framework on five benchmark polyp segmentation datasets: Kvasir-SEG \cite{jha2019kvasir}, CVC-ClinicDB \cite{bernal2015wm}, ETIS \cite{silva2014toward}, CVC-ColonDB \cite{tajbakhsh2015automated}, and EndoScene \cite{vazquez2017benchmark}. We use 1,450 images (900 from Kvasir-SEG and 550 from CVC-ClinicDB) for training, and the remaining 100 and 62 images for testing, following \cite{fan2020pranet}, thereby ensuring a fair comparison with SOTA models. To comprehensively assess the generalization capability, we further evaluate it on three additional datasets: CVC-ColonDB (380 images), ETIS (196 images), and CVC-300 (test set of EndoScene) (60 images).
\\
\textbf{Performance Metrics:} To ensure a fair comparison, we assess the performance of our approach as well as existing methods using six benchmark evaluation metrics such as mean Dice (mDice), mean IoU (mIoU), F-measure ($F_{\beta}^{w}$), S-measure ($S_{\alpha}$), E-measure ($E_{\phi}^{max}$), and mean absolute error ($\mathcal{M}$), as adopted in \cite{fan2020pranet}.\\ 
\textbf{Implementation Details: }
Our model is implemented in the PyTorch 2.7.1 framework and trained on an NVIDIA Tesla V100 GPU (32GB) with CUDA 11.8. All input images are uniformly resized to $352\times352$ pixels, and different data augmentation strategies, including multi-scale scaling with factors $\{0.75, 1.0, 1.25\}$ and random vertical/horizontal flipping, are adopted. The model training is performed for 100 epochs using the Adam optimizer with a batch size of 6 and an initial learning rate of $1\times10^{-4}$.

\vspace{-0.3cm}
\subsection{Quantitative Comparison}
We evaluate the proposed Lite-$\pi$ framework with three lightweight baselines: U-Net \cite{ronneberger2015u}, U-Net++ \cite{zhou2018unet++}, and PraNet \cite{fan2020pranet}. As shown in Table~\ref{tab:combined_seen}, Lite-$\pi$ consistently improves segmentation performance with minimal increase in parameters and FLOPs. On \textbf{seen datasets}, 
Lite-$\pi$ boosts mDice and mIoU across all baselines, achieving gains of +6-14\% on CVC-ClinicDB and +3-5\% on Kvasir-SEG, while reducing MAE. On \textbf{unseen datasets}, it yields strong generalization performance gains, improving mDice scores by nearly +15\% for baselines U-Net and U-Net++. Compared to Polyp-DiFoM \cite{agnihotri2026sam}, it delivers superior performance across all five datasets. For a fair comparison, the results of Polyp-DiFoM \cite{agnihotri2026sam} have been reproduced under the 3-FM configuration, which incorporates the same three FMs. 
Despite having significantly fewer parameters and FLOPs, Lite-$\pi$ enables lightweight baselines to achieve performance on par with recent FM-driven methods such as SAM-Mamba \cite{dutta2025sam} and SAM-MaGuP \cite{dutta2025ma}, which exceed 100M parameters. Additionally, our framework achieves an average latency of 17.5 ms and an average throughput of 62.6 FPS, with peak GPU memory usage of 326 MB, thereby establishing a new benchmark for real-time polyp segmentation. All methods are evaluated under the same dataset split, pre-processing pipeline, input resolution, and training settings for a fair comparison. Performance gains are robust, reproducible, and not dataset-specific

\begin{table*}[t]
\centering
\caption{Quantitative comparison of baselines with Lite-$\pi$ against SoTA methods on unseen datasets.}
\label{tab:combined_unseen}
\footnotesize
\renewcommand{\arraystretch}{0.85}
\setlength{\tabcolsep}{3pt}
\resizebox{\textwidth}{!}{%
\begin{tabular}{l|cccccc|cccccc|cccccc}
\toprule
\toprule
\multirow{2}{*}{Methods} 
& \multicolumn{6}{c|}{CVC-300 (Unseen)} 
& \multicolumn{6}{c|}{CVC-ColonDB (Unseen)} 
& \multicolumn{6}{c}{ETIS (Unseen)} \\
\cmidrule(lr){2-7}\cmidrule(lr){8-13}\cmidrule(lr){14-19}
& mDice$\uparrow$ & mIoU$\uparrow$ & $F_{\beta}^{w}$$\uparrow$ & $S_{\alpha}$$\uparrow$ & $E_{\phi}^{\max}$$\uparrow$ & $\mathcal{M}$$\downarrow$
& mDice$\uparrow$ & mIoU$\uparrow$ & $F_{\beta}^{w}$$\uparrow$ & $S_{\alpha}$$\uparrow$ & $E_{\phi}^{\max}$$\uparrow$ & $\mathcal{M}$$\downarrow$
& mDice$\uparrow$ & mIoU$\uparrow$ & $F_{\beta}^{w}$$\uparrow$ & $S_{\alpha}$$\uparrow$ & $E_{\phi}^{\max}$$\uparrow$ & $\mathcal{M}$$\downarrow$ \\
\midrule

\multicolumn{19}{c}{\textcolor{gray}{\textit{\textbf{State-of-the-art methods}}}} \\
SANet \cite{wei2021shallow}        & 88.8 & 81.5 & 85.9 & 92.8 & 97.2 & 0.8 & 75.3 & 67.0 & 72.6 & 83.7 & 87.8 & 4.3 & 75.0 & 65.4 & 68.5 & 84.9 & 89.7 & 1.5 \\
MSNet \cite{zhao2021automatic}     & 86.9 & 80.7 & 84.9 & 92.5 & 94.3 & 1.0 & 75.5 & 67.8 & 73.7 & 83.6 & 88.3 & 4.1 & 71.9 & 66.4 & 67.8 & 84.0 & 83.0 & 2.0 \\
Polyp-PVT \cite{Dong2023}           & 90.0 & 83.3 & 88.4 & 93.5 & 97.3 & 0.7 & 80.8 & 72.7 & 79.5 & 86.5 & 91.3 & 3.1 & 78.7 & 70.6 & 75.0 & 87.1 & 90.6 & 1.3 \\
PVT-Cascade \cite{rahman2023medical} & 89.2 & 82.4 & 87.3 & 93.2 & 95.9 & 0.9 & 78.1 & 71.0 & 77.9 & 85.5 & 89.6 & 3.1 & 78.6 & 71.2 & 75.9 & 87.2 & 89.6 & 1.3 \\
CFA-Net \cite{zhou2023cross}         & 89.3 & 82.7 & 93.8 & 87.5 & 97.8 & 0.8 & 74.3 & 66.5 & 72.8 & 83.5 & 89.8 & 3.9 & 73.2 & 65.5 & 69.3 & 84.5 & 89.2 & 1.4 \\
CTNet \cite{xiao2024ctnet}           & 90.8 & 84.4 & 89.4 & 97.5 & 97.5 & 0.6 & 81.3 & 73.4 & 80.1 & 87.4 & 91.5 & 2.7 & 81.0 & 73.4 & 77.6 & 88.6 & 91.3 & 1.4 \\
MEGANet \cite{bui2024meganet}        & 89.9 & 83.4 & 88.2 & 93.5 & 96.9 & 0.7 & 79.3 & 71.4 & 77.9 & 85.4 & 89.5 & 4.0 & 73.9 & 66.5 & 70.2 & 83.6 & 85.8 & 3.7 \\
SAM-Mamba \cite{dutta2025sam}        & 92.0 & 86.1 & 88.8 & 94.6 & 98.1 & 0.6 & 85.3 & 77.1 & 85.6 & 89.8 & 93.3 & 1.7 & 84.8 & 78.2 & 85.5 & 91.6 & 93.3 & 1.0 \\
SAM-MaGuP \cite{dutta2025ma}       & 92.7 & 88.0 & 90.1 & - & - & 0.5 & 85.9 & 78.5  & 86.2 & - & -  & 1.7 & 85.4 & 78.9 & 86.2 & - & - & 1.0\\
\midrule
\midrule

\multicolumn{19}{c}{\textcolor{gray}{\textit{\textbf{Lightweight baselines}}}} \\
U-Net \cite{ronneberger2015u}        & 71.0 & 62.7 & 68.4 & 84.3 & 87.6 & 2.2 & 51.2 & 44.4 & 49.8 & 71.2 & 77.6 & 6.1 & 39.8 & 33.5 & 36.6 & 68.4 & 74.0 & 3.6 \\
U-Net++ \cite{zhou2018unet++}        & 70.7 & 62.4 & 68.7 & 83.9 & 89.8 & 1.8 & 48.3 & 41.0 & 46.7 & 69.1 & 76.0 & 6.4 & 40.1 & 34.4 & 39.0 & 68.3 & 77.6 & 3.5 \\
PraNet \cite{fan2020pranet}          & 87.1 & 79.7 & 84.3 & 92.5 & 97.2 & 1.0 & 70.9 & 64.0 & 69.6 & 81.9 & 86.9 & 4.5 & 62.8 & 56.7 & 60.0 & 79.4 & 84.1 & 3.1 \\
\midrule

\multicolumn{19}{c}{\textcolor{gray}{\textit{\textbf{Lightweight baselines with Polyp-DiFoM (WACV'26)  \cite{agnihotri2026sam}}}}} \\
 Polyp-DiFoM (U-Net) & 82.3 & 73.3  & 74.7  & 89.4  & 86.9   & 1.6   
  & 68.3   & 57.4  & 60.9  & 81.4  & 77.1   & 4.7   
  & 54.3  & 42.4 & 47.1  & 76.6  & 70.1  & 2.6 \\
PolypDiFoM (U-Net++)   & 77.8  & 70.9  & 75.9  & 92.7   & 90.1  & 1.5   
  & 67.7  & 54.1  & 64.0 & 87.1 & 80.0 & 5.1
  & 51.8  & 42.7  & 49.0 & 80.1  & 72.0   & 3.2 \\
PolypDiFoM (PraNet)      & 87.4  & 79.9  & 87.1  & 95.9  & 97.9   & 0.8  
  & 74.0  & 64.3  & 73.0  & 88.4  & 85.9 & 3.9  
  & 71.5  & 58.2  & 65.6  & 87.0   & 83.5   & 2.2 \\
\bottomrule
\multicolumn{19}{c}{\textcolor{gray}{\textit{\textbf{Lightweight baselines empowered with Lite-$\pi$  (Ours)}}}} \\
Lite-$\pi$ (U-Net)        & 85.3  & 76.5   & 85.2  & 87.2 & 93.7  & 1.2  & 71.4  & 60.8  & 71.3 & 83.2  & 85.2  & 4.4  & 55.6  & 42.9 & 51.9 & 76.4 & 71.7 & 2.9 \\
& \textcolor{blue}{(+14.3)} & \textcolor{blue}{(+13.8)} & \textcolor{blue}{(+16.8)} & \textcolor{blue}{(+2.9)} & \textcolor{blue}{(+6.1)} & \textcolor{red}{(-1.0)} 
& \textcolor{blue}{(+20.2)} & \textcolor{blue}{(+16.4)} & \textcolor{blue}{(+21.5)} & \textcolor{blue}{(+12.0)} & \textcolor{blue}{(+7.6)} & \textcolor{red}{(-1.7)} 
& \textcolor{blue}{(+15.8)} & \textcolor{blue}{(+9.4)} & \textcolor{blue}{(+15.3)} & \textcolor{blue}{(+8.0)} & (-2.3) & \textcolor{red}{(-0.7)} \\
Lite-$\pi$ (U-Net++)       & 83.5  & 74.8  & 83.3  & 89.9  & 92.1  & 1.4  & 72.7  & 61.6   & 72.5  & 87.4  & 86.3  & 4.7  & 55.1  & 44.7  & 54.7  & 76.8  & 72.9  & 3.2 \\
& \textcolor{blue}{(+12.8)} & \textcolor{blue}{(+12.4)} & \textcolor{blue}{(+14.6)} & \textcolor{blue}{(+6.0)} & \textcolor{blue}{(+2.3)} & \textcolor{red}{(-0.4)} 
& \textcolor{blue}{(+24.4)} & \textcolor{blue}{(+20.6)} & \textcolor{blue}{(+25.8)} & \textcolor{blue}{(+18.3)} & \textcolor{blue}{(+10.3)} & \textcolor{red}{(-1.7)} 
& \textcolor{blue}{(+15.0)} & \textcolor{blue}{(+10.3)} & \textcolor{blue}{(+15.7)} & \textcolor{blue}{(+8.5)} & (-4.7) & \textcolor{red}{(-0.3)} \\
Lite-$\pi$ (PraNet)       & 89.4  & 83.6  & 89.8    & 97.1  & 95.4  & 0.6  & 75.9  & 70.1  & 76.7  & 88.4  & 89.1 & 3.8  & 72.9  & 61.6  & 66.2  & 80.8 & 84.9  & 1.5  \\
& \textcolor{blue}{(+2.3)} & \textcolor{blue}{(+3.9)} & \textcolor{blue}{(+5.5)} & \textcolor{blue}{(+4.6)} & (-1.8) & \textcolor{red}{(-0.4)} 
& \textcolor{blue}{(+5.0)} & \textcolor{blue}{(+6.1)} & \textcolor{blue}{(+7.1)} & \textcolor{blue}{(+6.5)} & \textcolor{blue}{(+2.2)} & \textcolor{red}{(-0.7)} 
& \textcolor{blue}{(+10.1)} & \textcolor{blue}{(+4.9)} & \textcolor{blue}{(+6.2)} & \textcolor{blue}{(+1.4)} & \textcolor{blue}{(+0.8)} & \textcolor{red}{(-1.6)} \\
\bottomrule
\bottomrule
\end{tabular}
}\vspace{-0.5cm}
\end{table*}

\begin{figure}[t]
\centering

\begin{subfigure}[t]{\linewidth}
    \centering
    \includegraphics[height=4.5cm,width=0.75\linewidth]{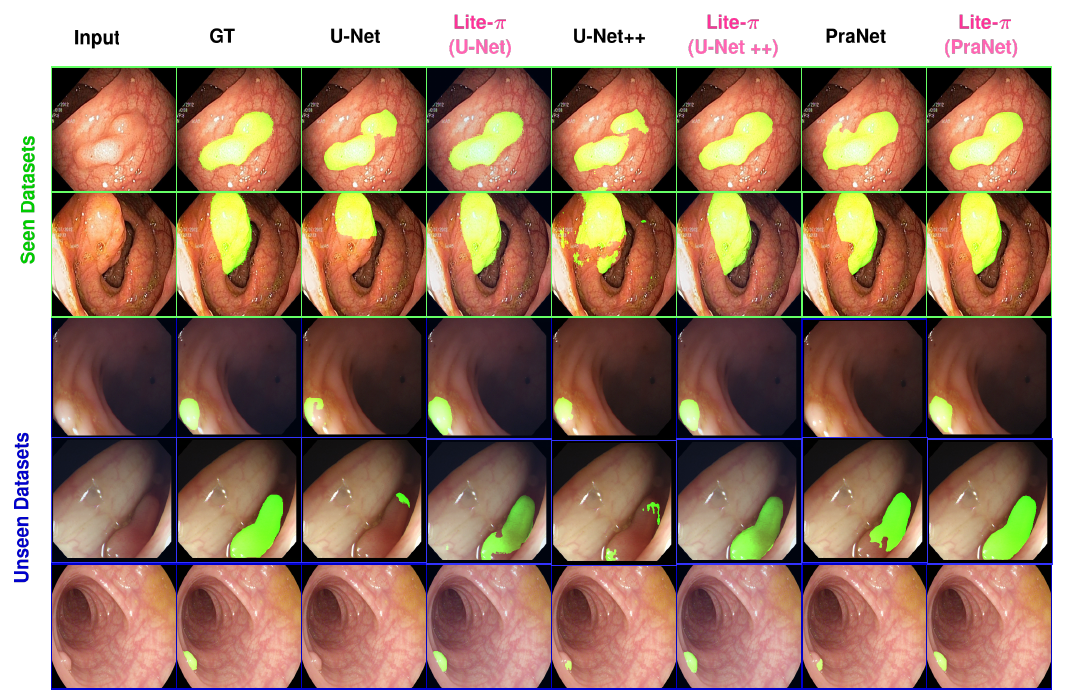}
    \caption{Qualitative comparison on seen and unseen datasets, demonstrating the ability of Lite-$\pi$ to accurately segment polyps of varying appearances.}
    \label{fig_qual_a}
\end{subfigure}

\vspace{0.2cm}

\begin{subfigure}[t]{\linewidth}
    \centering
    \includegraphics[height=2.5cm,width=7.5cm]{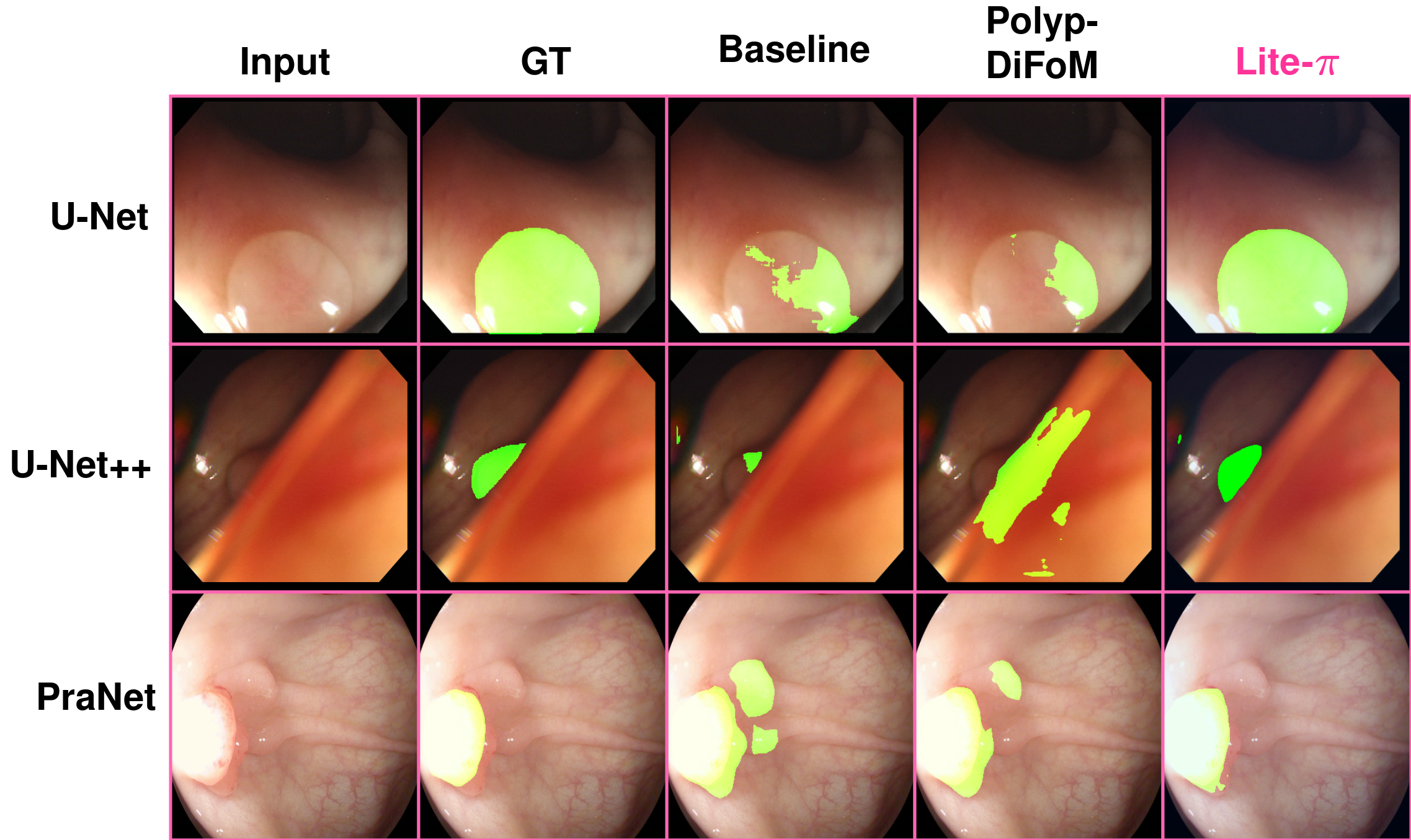}
    \caption{Comparison with Polyp-DiFoM \cite{agnihotri2026sam}, showing Lite-$\pi$'s superior segmentation results in complex polyp structures.}
    \label{fig_qual_b}
\end{subfigure}

\vspace{-0.2cm}
\caption{Qualitative results with a comparison against Polyp-DiFoM \cite{agnihotri2026sam}.}
\label{fig_qual}\vspace{-0.5cm}
\end{figure}

\subsection{Qualitative Comparison}
Fig. \ref{fig_qual_a} shows Lite-$\pi$ consistently achieves more accurate segmentation on both seen and unseen datasets, particularly in challenging scenarios where baseline models struggle, such as tiny polyps, weak boundaries, and low-contrast regions. U-Net produces highly inconsistent results, often missing polyps entirely or yielding false positives, whereas Lite-$\pi$ refines them and obtains better predictions consistently. Comparison with Polyp-DiFoM \cite{agnihotri2026sam} highlights significant improvements in segmentation results with all baselines (Fig. \ref{fig_qual_b}). While Polyp-DiFoM improves baseline predictions, it still fails to accurately detect polyps in several cases. These results demonstrate the significance of the UFMI module in enhancing the segmentation capability of lightweight baselines.
\vspace{-0.3cm}
\subsection{Ablation Study}
UFMI is designed as a unified module with tightly integrated components. Therefore, we assess the benefit of combining multiple FMs by progressively incorporating three FMs (SAM, DINOv2, and OneFormer) through ablation experiments rather than analyzing individual components. Table \ref{tab:ablation} shows that the best performance is obtained with the 3-FM configuration for all three baselines. In contrast to prior approaches, which exhibit performance saturation with progressive FM addition, our Lite-$\pi$ framework supports scalable multi-FM integration without significant computational burden.
\begin{table}[t]
\small
\centering
\renewcommand{\arraystretch}{0.45}
\setlength{\tabcolsep}{3pt}
\caption{Ablation study showing the effectiveness of incorporating multiple FMs into Lite-$\pi$.}
\vspace{0.0cm}
\resizebox{0.8\columnwidth}{!}{
\begin{tabular}{c|ccc|cc|ccc}
\toprule
\textbf{Baseline} & \textbf{SAM} & \textbf{DINOv2} & \textbf{OneFormer} 
& \bf \small Kvasir & \bf \small CVC-ClinicDB 
& \bf \small CVC-300 & \bf \small CVC-ColonDB & \bf \small ETIS \\
\midrule

\multirow{3}{*}{U-Net}
& \checkmark & - & - & 81.9 & 89.1 & 75.6 & 60.6 & 45.0 \\
& \checkmark & \checkmark & - & 85.7 & 92.1 & 80.7 & 67.7 & 50.9 \\
& \checkmark & \checkmark & \checkmark & \textbf{87.4} & \textbf{94.9} & \textbf{85.3} & \textbf{71.4} & \textbf{55.6} \\

\midrule

\multirow{3}{*}{U-Net++}
& \checkmark & - & - & 82.2 & 90.1 & 73.9 & 60.1 & 46.0 \\
& \checkmark & \checkmark & - & 84.6 & 92.1 & 81.3 & 67.1 & 52.1 \\
& \checkmark & \checkmark & \checkmark & \textbf{87.6} & \textbf{93.3} & \textbf{83.5} & \textbf{72.7} & \textbf{55.1} \\

\midrule

\multirow{3}{*}{PraNet}
& \checkmark & - & - & 89.5 & 93.4 & 86.7 & 74.6 & 69.2 \\
& \checkmark & \checkmark & - & 91.0 & 94.6 & 87.9 & \textbf{76.1} & \textbf{73.7} \\
& \checkmark & \checkmark & \checkmark & \textbf{92.9} & \textbf{95.4} & \textbf{89.4} & 75.9 & 72.9 \\

\bottomrule
\end{tabular}
}\vspace{-0.65cm}
\label{tab:ablation}
\end{table}
\section{Conclusion}\vspace{-0.35cm}
We present Lite-$\pi$, a new induction-based approach that effectively transfers structural and semantic cues from three FMs into a lightweight segmentation baseline. By synthesizing prototype representations and aligning them with the corresponding FMs through a reconstruction objective, followed by transformer- and Mamba-based feature refinement, our method facilitates efficient learning of high-level contextual and boundary-aware cues without incurring additional complexity. Extensive experiments across five polyp segmentation benchmarks demonstrate that Lite-$\pi$ consistently enhances segmentation accuracy, robustness, and generalization while maintaining low computational overhead. Lite-$\pi$ offers a practical and scalable solution for real-time clinical deployment where large foundation models are impractical. 
    

\begin{credits}
 \subsubsection{\ackname} This work is supported by the Anusandhan National Research Foundation (ANRF), Govt. of India (grant number ANRF/ARGM/2025/002890/TS and CRG/2023/007397 ).
\subsubsection{\discintname}
The authors have no competing interests to declare that are relevant to the content of this article.
\end{credits}

%
%
%
\bibliographystyle{splncs04}
\bibliography{refs}
%




\end{document}